\let\origvec\vec
\documentclass[runningheads]{llncs}
\usepackage{graphicx}

\let\vec\origvec
\usepackage{amsmath}
\usepackage{mathabx}
\usepackage{mathpazo}
\usepackage{booktabs}
\usepackage{multirow}
\usepackage{array}
\usepackage{subfigure}
\usepackage{color}
\usepackage{arydshln}
\usepackage{bm}
\usepackage{makecell}
\usepackage{cleveref}
\usepackage{hyperref}
\usepackage[super]{nth}
\begin{document}
\title{Brain Tumor Segmentation and Tractographic Feature Extraction from Structural MR Images for Overall Survival Prediction}
\titlerunning{Brain Tumor Segmentation and Overall Survival Prediction}
\author{Po-Yu Kao\inst{1} \and Thuyen Ngo\inst{1} \and Angela Zhang\inst{1} \and \\ Jefferson W. Chen\inst{2} \and 
B.S. Manjunath\inst{1}}

\authorrunning{P. Kao et al.}
\institute{Vision Research Lab, University of California, Santa Barbara, CA, USA\\
\email{\{poyu\_kao, manj\}@ece.ucsb.edu}\\
\and
UC Irvine Health, University of California, Irvine, CA, USA\\
}
\maketitle
\begin{abstract}
This paper introduces a novel methodology to integrate human brain connectomics and parcellation for brain tumor segmentation and survival prediction. For segmentation, we utilize an existing brain parcellation atlas in the MNI152 1mm space and map this parcellation to each individual subject data. We use deep neural network architectures together with hard negative mining to achieve the final voxel level classification.  For survival prediction, we present a new method for combining features from connectomics data, brain parcellation information, and the brain tumor mask. We leverage the average connectome information from the Human Connectome Project and map each subject brain volume onto this common connectome space. From this, we compute tractographic features that describe potential neural disruptions due to the brain tumor. These features are then used to predict the overall survival of the subjects. The main novelty in the proposed methods is the use of normalized brain parcellation data and tractography data from the human connectome project for analyzing MR images for segmentation and survival prediction. Experimental results are reported on the BraTS2018 dataset.
\keywords{Brain Tumor Segmentation \and Brain Parcellation \and Group Normalization \and Hard Negative Mining \and Ensemble Modeling \and Overall Survival Prediction \and Tractographic Feature}
\end{abstract}
\section{Introduction}
Glioblastomas, or Gliomas, are one of the most common types of brain tumor. They have a highly heterogeneous appearance and shape and may happen at any location in the brain. High-grade glioma (HGG) is one of the most aggressive types of brain tumor with median survival of 15 months \cite{thakkar2014epidemiologic}. There is a significant amount of recent work on brain tumor segmentation and survival prediction. Kamnitsas et al. \cite{kamnitsas2017ensembles} integrate seven different 3D neural network models with different parameters and average the output probability maps from each model to obtain the final brain tumor mask. Wang  et al. \cite{wang2017automatic} design a hierarchical pipeline to segment the different types of tumor compartments using anisotropic convolutional neural networks. The network architecture of  Isensee et al. \cite{isensee2017brain} is derived from a 3D U-Net with additional residual connections on context pathway and additional multi-scale aggregation on localization pathways, using the Dice loss in the training phase to circumvent class imbalance. For the brain tumor segmentation task, we propose a methodology to integrate multiple DeepMedics \cite{kamnitsas2017efficient} and patch-based 3D U-Nets adjusted from \cite{cciccek20163d} with different parameters and different training strategies in order to get a robust brain tumor segmentation from multi-modal structural MR images. We also utilize the existing brain parcellation to bring location information to the patch-based neural networks. In order to increase the diversity of our ensemble, 3D U-Nets with dice loss and cross-entropy loss are included. The final segmentation mask of the brain tumor is calculated by taking the average of the output probability maps from each model in our ensemble. 

For the overall survival (OS) prediction task, Shboul et al. \cite{shboul2017glioblastoma} extract 40 features from the predicted brain tumor mask and use a random forest regression to predict the glioma patient's OS. Jungo et al. \cite{alain2017towards} extract four features from each subject and use a support vector machine (SVM) with radial basis function (RBF) kernel to classify glioma patients into three different OS groups. In this paper, we propose a novel method to extract the tractographic features from the lesion regions on structural MR images via an average diffusion MR image which is from a total of 1021 HCP subjects \cite{van2013wu} (Q1-Q4, 2017). We then use these tractographic features to predict the patient's OS with a SVM classifier with linear kernel.

\section{Glioma Segmentation}
\label{sec:seg}
\subsection{Materials}
The Brain Tumor Segmentation (BraTS) 2018 dataset  \cite{bakas2017lgg,bakas2017gbm,bakas2017advancing,menze2015multimodal} provides 285 training subjects with four different types of MR images (MR-T1, MR-T1ce, MR-T2 and MR-FLAIR) and expert-labeled ground-truth of lesions, including necrosis \& non-enhancing tumor, edema, and enhancing tumor. The dataset consists of 66 validation subjects and 191 test subjects with four different types of MR images. These MR images are co-registered to the same anatomical template, interpolated to the same resolution ($1mm^3$) and skull-stripped. For each subject, a standard z-score normalization is applied within the brain region as our pre-processing step for brain tumor segmentaion. 
\subsection{Brain Parcellation Atlas as a Prior for Tumor Segmentation}
\begin{figure}[htbp]
\includegraphics[width=\textwidth]{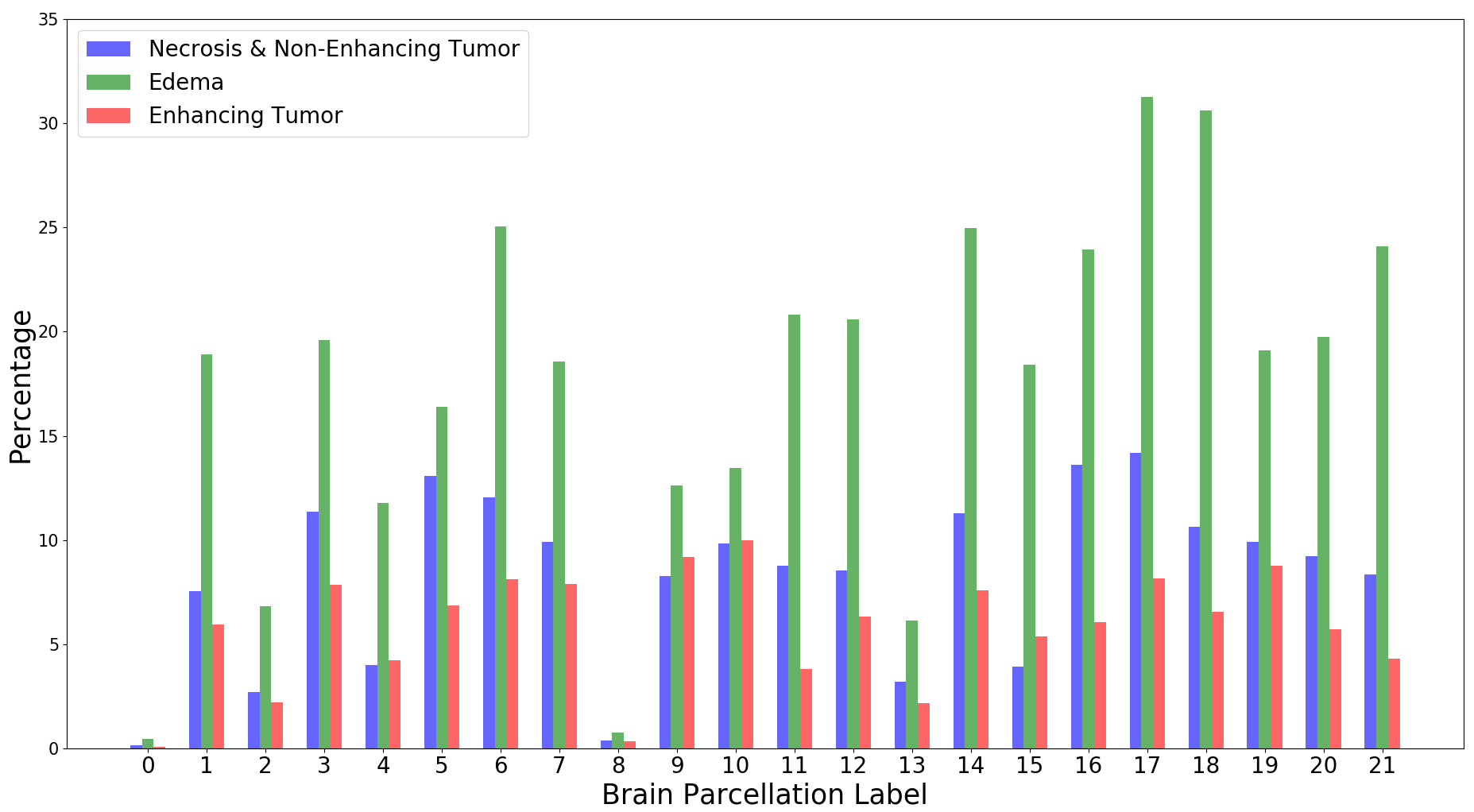}
\caption{The percent of brain lesion types observed in different parcellation regions of the Harvard-Oxford subcortical atlas \cite{desikan2006automated}. The $x$-axis indicates the parcellation label. Regions not covered by the Harvard-Oxford subcortical atlas are in label 0.}
\label{fig:lesionProb}
\end{figure}

\begin{figure}[htbp]
\includegraphics[width=\textwidth]{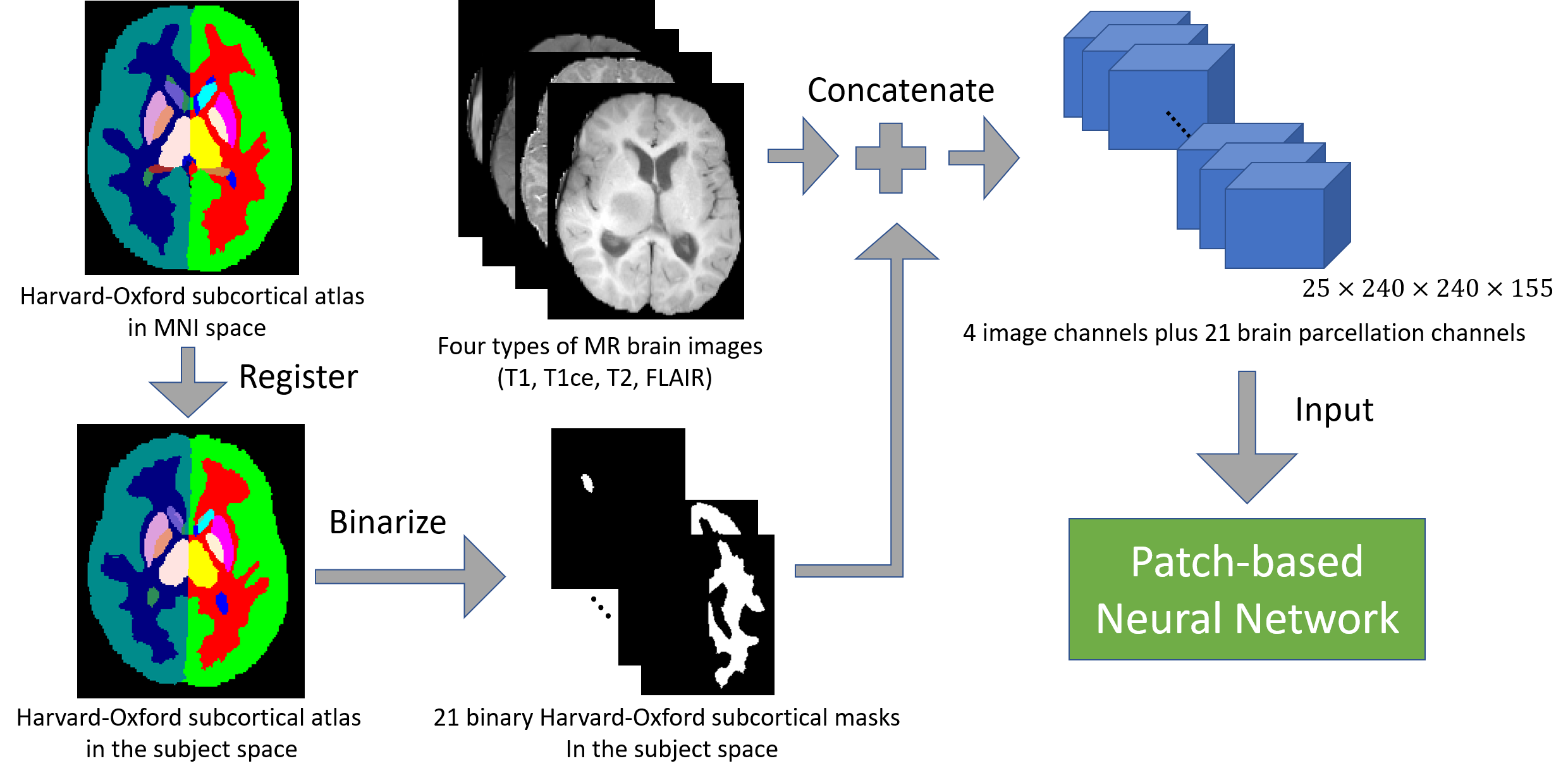}
\caption{Incorporating brain parcellation atlas into a patch-based neural network. First, Harvard-Oxford subcortical atlas is registered to the subject space, and the parcellation label is binarized into a 21-dimension vector. This vector is concatenated with the original MR images as input to a patch-based neural network.} 
\label{fig:bp2cnn}
\end{figure}

Current state-of-the-art deep network architectures \cite{isensee2017brain,kamnitsas2017ensembles,wang2017automatic} for brain tumor segmentation do not consider location information. However, from Figure \ref{fig:lesionProb}, it is clear that the lesions are not uniformly distributed in different brain regions. This distribution is computed by dividing the total volume of the lesions by the total volume of the corresponding brain parcellation region. Our proposed method (Figure \ref{fig:bp2cnn}) explicitly includes the location information as input into a patch-based neural network. First, we register the brain parcellation atlas to the subject space using FLIRT\cite{jenkinson2001global} from FSL. This registration enables associating each subject voxel with a structure label indicating the voxel location normalized across all subjects. Thus, the input to the neural network will include both the image data and the corresponding parcellation labels.

\subsection{Network Architecture and Training}

We integrate multiple state-of-the-art neural networks in our ensemble \footnote[1]{The ensemble is publicly available at \url{https://hub.docker.com/r/pykao/brats2018/}} for robustness. Our ensemble combines 26 neural networks adapted from \cite{cciccek20163d,kamnitsas2017efficient}. The detailed network architecture and training method for each model is shown in Table \ref{tab:26_models}. Each 3D U-Net uses group normalization \cite{wu2018group} and each DeepMedic uses batch normalization in our ensemble. We utilize a hard negative mining strategy to solve the class imbalance problem while we train a 3D U-Net with cross-entropy loss. Finally, we take the average of the output probability maps from each neural network and get the final brain tumor segmentation. The average training time for each DeepMedic is approximately 3 hours and for each 3D U-Net is approximately 12 hours, and the average testing time for a subject is approximately 20 minutes on a NVIDIA GTX Titan X and a Intel Xeon CPU E5-2696 v4 @ 2.20GHz.

\begin{table}[htbp]
\caption{The network architecture of 26 models in our ensemble. Models \#1 to \#6, \#18 and \# 19 have the same architecture but different initializations, and models \#21 to \#26 have the same architecture but different initializations. DeepMedic uses batch normalization and 3D U-Net uses group normalization. DeepMedic and models \#23 to \#26 are trained with the cross-entropy loss. The batch size for \#3 to \#19 is 50 and for 3D U-Net is 2. The input patch size for model \#1 to \#17 is $25\times25\times25$ and for 3D U-Net is $128\times128\times128$. 3D U-Nets and DeepMedics without additional brain parcellation channels are trained with 300 epochs, DeepMedic with additional brain parcellation channels are trained with 500 epochs, and models \#18 and \#19 are trained with 600 epochs. Adam \cite{kingma2014adam} is used with 0.001 learning rate in the optimization step for all models. (\# : model number, BP: input Harvard-Oxford subcortical atlas with MR images to the model, Aug.: data augmentations including random flipping in $x$-, $y$- and $z$-dimension.)}
\label{tab:26_models}
\centering
\begin{tabular}{ |m{0.5cm}|m{0.3cm}|m{0.4cm}|m{0.7cm}|m{5.5cm}|| }
\hline
& \# & BP & Aug. & Note \\
\hline
\hline
\parbox[l]{2mm}{\multirow{19}{*}{\rotatebox[origin=c]{90}{\centering \Large{DeepMedic}}}} & 1 & & & \multirow{2}{*}{Batch size: 36} \\\cline{2-4}
& 2  & & ${\surd}$& \\\cline{2-5}
& 3 &  & & \\\cline{2-5}
&4 & ${\surd}$& &\\\cline{2-5}
&5  &  & ${\surd}$ &\\\cline{2-5}
&6  & ${\surd}$ & ${\surd}$ &\\\cline{2-5}
&7  &  & ${\surd}$ & \multirow{2}{*}{1.5 times 3D convolutional kernels} \\\cline{2-4}
&8  & ${\surd}$ & ${\surd}$ & \\\cline{2-5}
&9  &  & &\multirow{3}{*}{Double 3D convolutional kernels}\\\cline{2-4}
&10 &  & ${\surd}$ &\\\cline{2-4}
&11 & ${\surd}$ & ${\surd}$  &\\\cline{2-5}
&12 &  & & \multirow{3}{*}{2.5 times 3D convolutional kernels}\\\cline{2-4}
&13 &  & ${\surd}$ &\\\cline{2-4}
&14 & ${\surd}$ & ${\surd}$ &\\ \cline{2-5}
&15 &  & & \multirow{3}{*}{Triple 3D convolutional kernels}\\\cline{2-4}
&16 &  & ${\surd}$ & \\\cline{2-4}
&17 & ${\surd}$ & ${\surd}$ & \\\cline{2-5}
&18 &  & ${\surd}$ & Input patch size: $22\times22\times22$ \\\cline{2-5}
&19 &  & ${\surd}$ & Input patch size: $28\times28\times28$\\
\hline
\hline
\parbox[l]{2mm}{\multirow{7}{*}{\rotatebox[origin=c]{90}{\centering \Large{3D U-Net}}}} & 20 & & & From \cite{isensee2017brain} with Dice loss\\\cline{2-5}
& 21 &  & & \multirow{2}{*}{Dice loss}\\\cline{2-4}
& 22 & ${\surd}$ & & \\\cline{2-5}
& 23 &  & & \multirow{2}{*}{hard negative mining within one batch}\\\cline{2-4}
& 24 & ${\surd}$ & & \\\cline{2-5}
& 25 &   & & \multirow{2}{*}{hard negative mining within one image}\\\cline{2-4}
& 26 & ${\surd}$ & & \\

\hline
\end{tabular}
\end{table}
\subsubsection{Group Normalization}
The deep network architectures used for segmentation are computationally demanding. For the 3D U-Nets, our GPU resources enable us to use only 2 samples (of dimensions $128\times128\times128$ voxels) per iteration. With this small batch size of 2 samples, batch statistics collected during conventional batch normalization method \cite{ioffe2015batch} are unstable and thus not suitable for training. In batch normalization, statistics are computed for each feature dimension. Recently Wu et al. \cite{wu2018group} propose to group several feature dimensions together while computing batch statistics. This so-called group normalization helps to stabilize the computed statistics. In our implementation, the number of groups is set to 4. 

\subsubsection{Hard Negative Mining}
We train a 3D U-Net with $128\times128\times128$ patches randomly cropped from the original data. With such large dimensions, the majority of voxels are not classified as lesion and the standard cross-entropy loss would encourage the model to favor the background class. To cope with this problem, we only select negative voxels with the largest losses (hard negative) to back-propagate the gradients. In our implementation, the number of selected negative voxels is at most three times the number of positive voxels. Hard negative mining not only improves the tumor segmentation performance of our model but also decreases its false positive rate.

\subsection{Experimental Results}

 We first examine the brain tumor segmentation performance using MR images and the Harvard-Oxford subcortical brain parcellation masks as input to DeepMedic and 3D U-Net. The quantitative results are shown in Table \ref{tab:bp_result}. This table demonstrates that adding brain parcellation masks as additional inputs to a patch-based neural network improves its performance. For segmentation of the enhancing tumor, whole tumor and tumor core, the average \textit{Hausdorff 95} scores for DeepMedic-based models improve from 5.205 to 3.922, from 11.536 to 8.507 and from 11.215 to 8.957, respectively. The average \textit{Dice} scores for models based on 3D U-Net also improve from 0.753 to 0.764, from 0.889 to 0.894 and from 0.766 to 0.775, respectively, for each of the three tumor compartments.

\begin{table}[htbp]
\caption{Quantitative results of the performance of adding additional brain parcellation masks with MR images to DeepMedic and 3D U-Net on the BraTS2018 validation dataset. Bold numbers highlight the improved results with additional brain parcellation masks. Models with BP use binary brain parcellation masks and MR images as input, while models without BP use only MR images as input. For comparison, each model without brain parcellation (BP) is paired with the same model using BP, the pair having the same parameters and weights initially. The scores for DeepMedic without BP is the average scores from model \#3, \#5, \#7, \#10, \#13 and \#16, and the scores for DeepMedic with BP is the average scores from model \#4, \#6, \#8, \#11, \#14 and \#17. The scores for 3D U-Net without BP is the average scores from model \#21, \#23 and \#25, and the scores for 3D U-Net with BP is the average scores from model \#22, \#24 and \#26. Tumor core (TC) is the union of necrosis \& non-enhancing tumor and enhancing tumor (ET). Whole tumor (WT) is the union of necrosis \& non-enhancing tumor, edema and enhancing tumor. Results are reported as mean.}
\label{tab:bp_result}
\centering
\begin{tabular}{ |m{5.5em} |m{4cm}|m{3em}|m{3em}|m{3em}| }
\hline
  {}   &  Description & ET & WT  & TC\\
\hline\hline
\multirow{4}{*}{$Dice$} 
& DeepMedic without BP  & 0.758  & 0.892 & 0.804  \\\cdashline{2-5}
& DeepMedic with BP     & \textbf{0.766}  & \textbf{0.894} & 0.804   \\\cline{2-5}
& 3D U-Net without BP  & 0.753 & 0.889 & 0.766\\\cdashline{2-5}
& 3D U-Net with BP  & \textbf{0.764} & \textbf{0.894} & \textbf{0.775}\\
\hline
\hline
\multirow{4}{*}{\shortstack{$\mathit{Hausdorff ~95}$\\{\tiny (in $mm$)}}}
& DeepMedic without BP  & 5.205  & 11.536 & 11.215  \\\cdashline{2-5}
& DeepMedic with BP     & \textbf{3.992}  & \textbf{8.507} & \textbf{8.957}   \\\cline{2-5}
& 3D U-Net without BP  & 4.851 & 5.337 & 10.550\\\cdashline{2-5}
& 3D U-Net with BP  & 5.216 & 5.544 & \textbf{10.442}\\
\hline
\end{tabular}
\end{table}

We then evaluate the brain tumor segmentation performance of our proposed ensemble on the BraTS2018 training, validation and test datasets. The quantitative results are shown in Table \ref{tab:ensemble_result}. This table shows the robustness of our ensemble on the brain tumor segmentation task. Our ensemble has consistent brain tumor segmentation performance on the BraTS2018 training, validation and test datasets. 

\begin{table}[htbp]
\caption{Quantitative results of the tumor segmentation performance of our ensemble on BraTS2018 training dataset with 5-fold cross-validation, validation dataset and test dataset. Tumor core (TC) is the union of necrosis \& non-enhancing tumor and enhancing tumor (ET). Whole tumor (WT) is the union of necrosis \& non-enhancing tumor, edema and enhancing tumor. Results are reported as mean.}
\label{tab:ensemble_result}
\centering
\begin{tabular}{ ||m{5.5em} |m{3.2cm}|m{3em}|m{3em}|m{3em}|| }
\hline
  {}   &  Dataset & ET & WT  & TC\\
\hline\hline
\multirow{3}{*}{$Dice$} 
& BraTS2018 training & 0.735 & 0.902& 0.813 \\\cline{2-5}
& BraTS2018 validation & 0.788 & 0.905& 0.813 \\\cline{2-5}
& BraTS2018 test & 0.749& 0.875& 0.793 \\
\hline
\hline
\multirow{3}{*}{\shortstack{$\mathit{Hausdorff ~95}$\\{\tiny (in $mm$)}}} 
& BraTS2018 training & 5.433 & 5.398 & 6.932 \\\cline{2-5}
& BraTS2018 validation &3.812 & 4.323 & 7.555 \\\cline{2-5}
& BraTS2018 test & 4.219 & 6.479 & 6.522 \\
\hline
\end{tabular}
\end{table}

\section{Overall Survival Prediction for Brain Tumor Patients}
\subsection{Material}
The BraTS2018 dataset also includes the age (in years), survival (in days) and resection status for each of 163 subjects in the training dataset, and 59 of them have the resection status of Gross Total Resection (GTR). The validation dataset has 53 subjects with the age (in years) and resection status, and 28 of them have the resection status of GTR. The test dataset has 131 subjects with the age (in years) and resection status, and 77 of them have the resection status of GTR. For this task, we only predict the overall survival (OS) for glioma patients with resection status of GTR.
\subsection{Methodology}
\begin{figure}
\includegraphics[width=\textwidth]{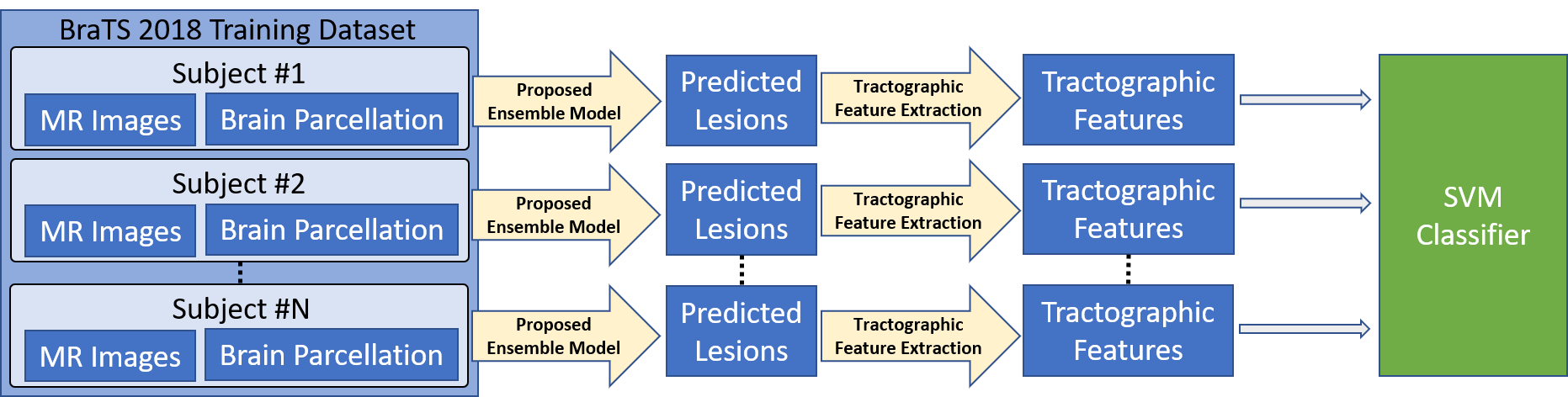}
\caption{Training pipeline for overall survival prediction.} \label{os_training}
\end{figure}

Our proposed training pipeline, shown in Figure \ref{os_training}, includes three stages: In the first stage, we use the proposed ensemble from the section \ref{sec:seg} to obtain the predicted tumor mask for each subject. In the second stage, We extract the tractographic features explained in section below from each subject. We then perform feature normalization and selection. In the final stage, we train a SVM classifier with linear kernel using the tractographic features extracted from the training subjects. We evaluate the overall survival classification performance of tractographic features on the BraTS2018 training dataset with the 1000-time repeated stratified 5-fold cross-validation, valdiation datset and test dataset.

\subsubsection*{Glioma Segmentation:}
To segment the glioma, we use the proposed ensemble in the previous section to obtain the prediction of three different types of tissue including necrosis \& non-enhancing tumor, edema, and enhancing tumor. 
\subsubsection*{Tractographic Feature Extraction from the Glioma Segmentation:}
After we obtain the predicted lesion mask, we extract the tractographic features from the whole tumor region which is the union of all different lesions for each subject. 

\begin{figure}
\includegraphics[width=\textwidth]{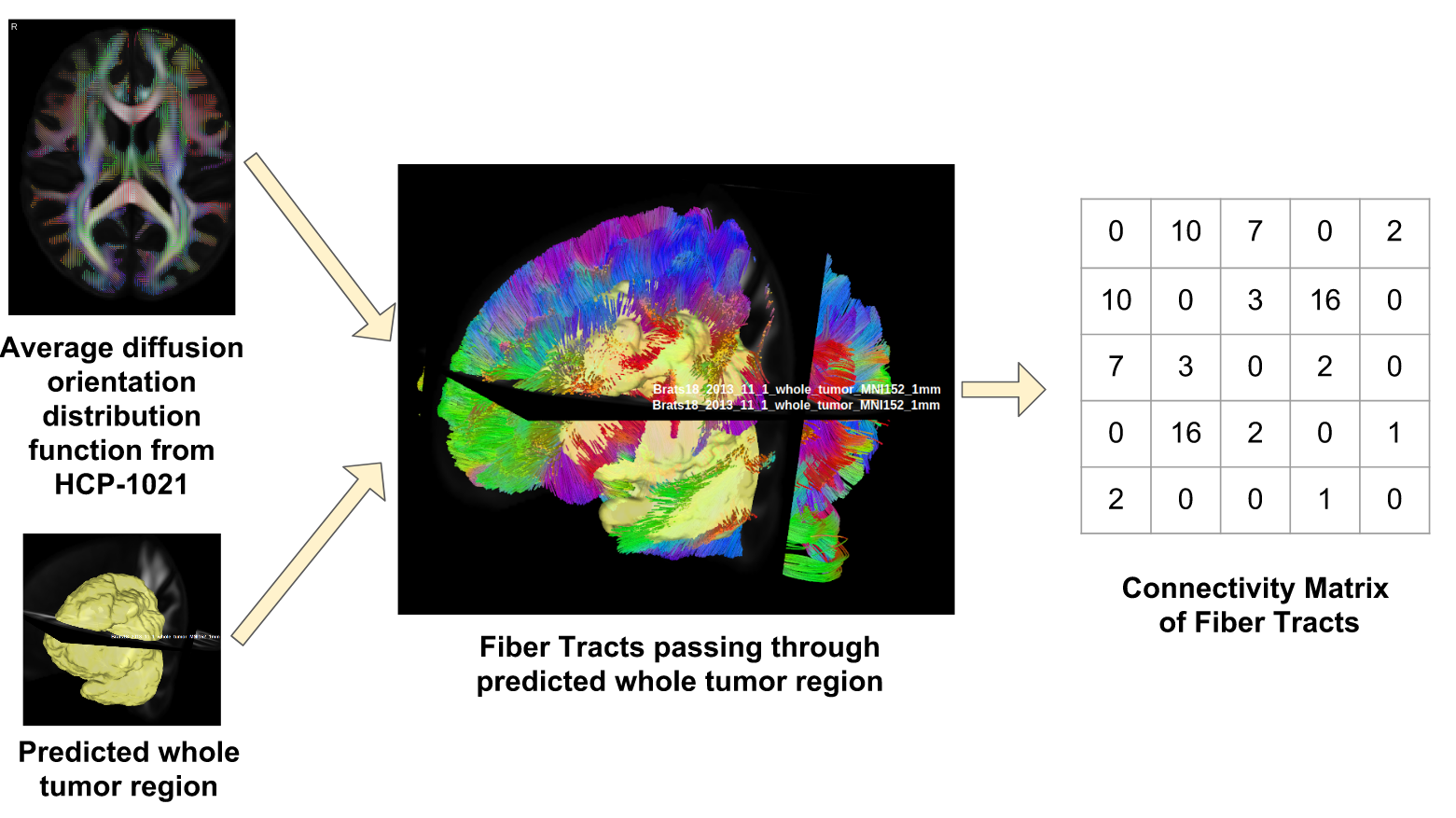}
\caption{Workflow for building a connectivity matrix for each subject. The fiber tracts are created by DSI Studio (\url{http://dsi-studio.labsolver.org/}), and ITK-SNAP \cite{py06nimg} is used for visualizing the 3D MR images and 3D labels.} \label{connectivity}
\end{figure}
\paragraph{Tractographic Features:}
Tractographic features describe the potentially damaged parcellation regions impacted by the brain tumor through fiber tracking. Figure \ref{connectivity} shows the workflow for building a connectivity matrix for each subject. First, the predicted whole tumor mask and the average diffusion orientation distribution function from HCP-1021, created by QSDR \cite{yeh2011ntu}, are obtained for each subject. FLIRT is used to map the whole tumor mask from subject space to MNI152 1mm space. Second, we use a deterministic diffusion fiber tracking method \cite{yeh2013deterministic} to create approximately 1,000,000 tracts from the whole tumor region. Finally, a structural brain atlas is used to create a connectivity matrix $\vec{W}_{ori}$ for each subject. This matrix contains information about whether a fiber connecting one region to another passed through or ended at those regions, as shown:

$\vec{W}_{ori}$ is a $N \times N$ matrix, and $N$ is the number of parcellation in a structural brain atlas.
\begin{equation}
\vec{W}_{ori} = 
\begin{bmatrix}
w_{ori,11}       & w_{ori,12} & \dots & w_{ori,1N} \\
w_{ori,21}       & w_{ori,22} & \dots & w_{ori,2N} \\
\vdots       & \vdots & \ddots& \vdots \\
w_{ori,N1}       & w_{ori,N2} & \dots & w_{ori,NN}
\end{bmatrix}
\end{equation}
If $w_{ij}$ is pass-type, it shows the number of tracts passing through region $j$ and region $i$. if $w_{ij}$ is end-type, it shows the number of tracts starting from a region $i$ and ending in a region $j$.
From the original connectivity matrix $\vec{W}_{ori}$, we create a normalized version $\vec{W}_{nrm}$ and a binarized version $\vec{W}_{bin}$.
\begin{equation}
    \vec{W}_{nrm} = \vec{W}_{ori} / max(\vec{W}_{ori})
\end{equation}
$/$ is the element-wise division operator, and $max(\vec{W}_{ori})$ is the maximum value of the original connectivity matrix $\vec{W}_{ori}$.
\begin{equation}
\vec{W}_{bin} = 
\begin{bmatrix}
w_{bin,11}       & w_{bin,12} & \dots & w_{bin,1N} \\
w_{bin,21}       & w_{bin,22} & \dots & w_{bin,2N} \\
\vdots       & \vdots & \ddots& \vdots \\
w_{bin,N1}       & w_{bin,N2} & \dots & w_{bin,NN}
\end{bmatrix}
\end{equation}
$w_{bin, ij} = 0$ if $w_{ori,ij} = 0$, and $w_{bin, ij} = 1$ if $w_{ori,ij} > 0$.
Then, we sum up each column in a connectivity matrix to form a unweighted tractographic feature vector. 
\begin{equation}
\vec{V} 
= \sum_{i=1}^{N} w_{ij} = 
\begin{bmatrix}
v_1, v_2, \dots, v_N
\end{bmatrix}
\end{equation}
Furthermore, we weight every element in the unweighted tractographic feature vector with respect to the ratio of the lesion in a brain parcellation region to the volume of this brain parcellation region. 
\begin{equation}
\vec{V}_{wei} 
= \vec{\alpha}\odot \vec{V}, \vec{\alpha} = 
\begin{bmatrix}
t_1/b_1, t_2/b_2, \dots, t_N/b_N
\end{bmatrix}
\end{equation}
$\odot$ is the element-wise multiplication operator, $t_i$ is the volume of the whole brain tumor in the $i$-th brain parcellation, and $b_i$ is the volume of the $i$-th brain parcellation. This vector $\vec{V}_{wei}$ is the tractographic feature extracted from brain tumor.  

In this paper, automated anatomical labeling (AAL) \cite{tzourio2002automated} is used for building the connectivity matrix. AAL has 116 brain parcellation regions, so the dimension of the connectivity matrix $\vec{W}$ is $116 \times 116$ and the dimension of each tractographic feature $\vec{V}_{wei}$ is $1 \times 116$. In the end, we extract six types of tractographic features for each subject. Six types of tractographic features are computed from: 1) the pass-type of the original connectivity matrix, 2) the pass-type of the normalized connectivity matrix, 3) the pass-type of the binarized connectivity matrix, 4) the end-type of the original connectivity matrix, 5) the end-type of the normalized connectivity matrix and 6) the end-type of the binarized connectivity matrix. 
\subsubsection*{Feature Normalization and Selection:}
First, we remove features with low variance between subjects, and then apply a standard z-score normalization on the remaining features. In the feature selection step, we combine recursive feature elimination with the 1000-time repeated stratified 5-fold cross-validation and a SVM classifier with linear kernel. These feature processing steps are implemented by using scikit-learn \cite{scikit-learn}.
\subsubsection*{Overall Survival Prediction:}
We first divide all 59 training subjects into three groups: long-survivors (e.g., \textgreater 15 months), short-survivors (e.g., \textless 10 months), and mid-survivors (e.g., between 10 and 15 months). Then, we train a SVM classifier with linear kernel on all training subjects with 1000-time repeated stratified 5-fold cross-validation in order to evaluate the performance of the proposed tractographic feature on overall survival prediction for brain tumor patients. We also evaluate the OS prediction performance of tractographic features on the BraTS2018 validation and test dataset. 

\subsection{Experimental Results}

In this task, we first examine the overall survival classification performance of our proposed tractographic feature compared to other types of features including age, volumetric features, spatial features, volumetric spatial features and morphological features. 

\paragraph{Volumetric Features:}
The volumetric features include the volume and the ratio of brain to the different types of lesions, as well as the tumor compartments. 19 volumetric features are extracted from each subject. 

\paragraph{Spatial Features:}
The spatial features describe the location of the tumor in the brain. The lesions are first registered to the MNI152 1mm space by using FLIRT, and then the centroids of whole tumor, tumor core and enhancing tumor are extracted as our spatial features. For each subject, we extract 9 spatial features. 

\paragraph{Volumetric Spatial Features:}
The volumetric spatial features describe the volume of different tumor lesions in different brain regions. First, the Harvard-Oxford subcortical structural atlas brain parcellation regions are registered to the subject space by using FLIRT. The volumes of different types of tumor lesions in each of parcellation regions, left brain region, middle brain region, right brain region and other brain region are extracted as volumetric spatial features. For each subject, we extract 78 volumetric spatial features.
\paragraph{Morphological Features:}
The morphological features include the length of the major axis of the lesion, the length of the minor axis of the lesion and the surface irregularity of the lesions. We extract 19 morphological features from each subject.

\begin{figure}[htbp]
\centering
\includegraphics[width=\textwidth]{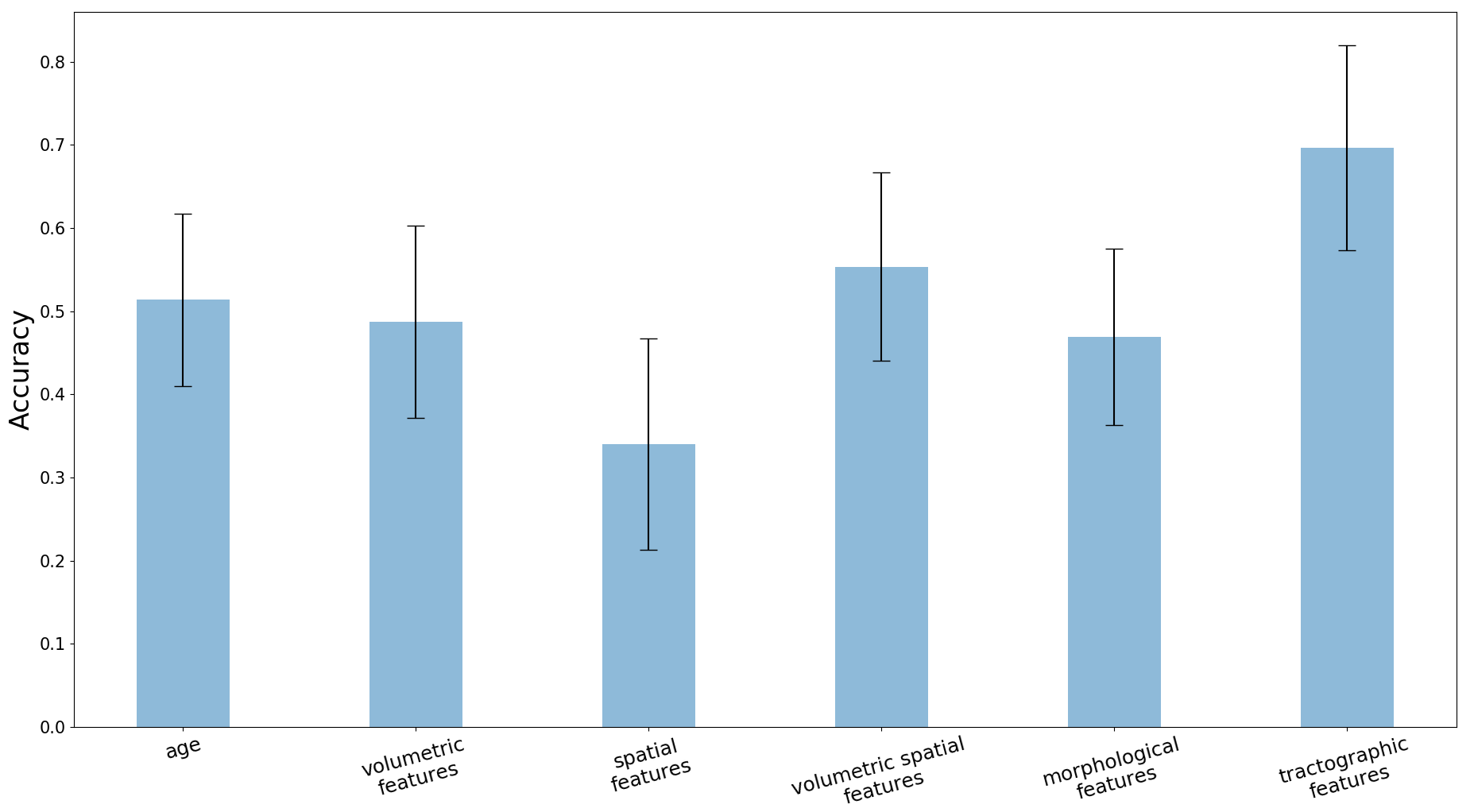}
\caption{Overall survival classification accuracy between different types of features on BraTS2018 training dataset. 1000-time repeated stratified 5-fold cross-validation is used to obtain the average classification accuracy.} \label{classification_accuracy_types}
\end{figure}

In the first experiment, the ground-truth lesion is used to extract different types of features, and the pass-type of the binarized connectivity matrix is built to compute the tractographic feature. Recursive feature elimination with cross-validation (RFECV) is used in the feature selection step to shrink the feature. A SVM classifier with linear kernel is trained with each feature type, and stratified 5-fold cross-validation is conducted 1000 times in order to achieve a reliable metric. The average and standard deviation of overall survival classification accuracy for different types of features on the BraTS2018 training dataset is shown in Figure \ref{classification_accuracy_types}. This figure demonstrates that the proposed tractographic features have the best overall survival classification performance compared to age, volumetric features, spatial features, volumetric spatial features and morphological features. Initial analysis based on feature selection indicate that 12 out of 116 AAL regions are more influential in affecting overall survival of the brain tumor patient. 

Next, the pass-type of the binarized connectivity matrix is built from the predicted lesion and the tractographic feature is computed from this connectivity matrix. The overall survival classification performance of this tractographic feature is compared with the tractographic feature from our first experiment. In this experiment, we follow the same feature selection method and training strategy, using the same SVM classifier with linear kernel. The average and standard deviation of overall survival classification accuracy on the BraTS2018 training dataset is reported in Table \ref{tab:classification_accuracy_lesions}. From this table, the average classification accuracy drops to 63 \%  when we use predicted lesions instead of ground-truth lesions to generate the tractographic features. This drop is likely caused by the imperfection of our tumor segmentation tool. 

\begin{table}[htbp]
    \centering
    \begin{tabular}{c|c}
    The source of tractographic features & Classification accuracy (mean$\pm$std) \\
    \hline
    Ground-truth Lesions     & 0.70 $\pm$ 0.12 \\
    \hline
    Predicted Lesions     &  0.63 $\pm$ 0.13
    \end{tabular}
    \caption{The overall survival classification performance of the proposed tractographic features from the ground-truth lesions and from the predicted lesions on the BraTS2018 training dataset with 1000-time repeated stratified 5-fold cross-validation.}
    \label{tab:classification_accuracy_lesions}
\end{table}

For the training data, the tractographic features are computed using the ground-truth whole tumor, and a linear SVM classifier trained on these features. We used stratified 5-fold cross validation on the training dataset, averaged over 1000 independent trials. The average OS classification accuracy using the tractographic features was 0.892 on the training set and 0.697 on the cross-validation set. However, when applied to the BraTS2018 validation and test datasets, the accuracy dropped to 0.357 and 0.416, respectively \cite{bakas2018identifying}. Note that for the validation and test data, there is no ground-truth segmentation available. So we first predicted the whole tumor and then the tractography features are extracted from these predicted tumors, followed by the OS classification using the previously trained linear SVM. We speculate that the automated segmentation to predict the whole tumor is one possible reason for the significant variation in performance between the training and validation/test data, in addition any data specific variations.

\section{Discussion}

For brain tumor segmentation, our proposed method, which combines the lesion occurrence probabilities in structural regions with MR images as inputs to a patch-based neural network, improves the patch-based neural network's performance. The proposed ensemble results in a more robust tumor segmentation. For overall survival prediction, the novel use of tractographic features appears to be promising for aiding brain tumor patients. To the best of our knowledge, this is the first paper to integrate brain parcellation and human brain connectomics for brain tumor segmentation and overall survival prediction. 

\subsubsection*{Acknowledgements.} This research was partially supported by a National Institutes of Health (NIH) award \# 5R01NS103774-02.
%

%
%
%
\bibliographystyle{splncs04}
\bibliography{references}
\end{document}